%% file: main.tex
\documentclass[indigo,fancycite]{meta}

\usepackage{xspace}
\usepackage{amsmath,amssymb}
\usepackage{tabularx,booktabs,array,colortbl}
\usepackage{float}
\usepackage{algorithm}
\usepackage{algorithmic}
\usepackage{mdframed}
\usepackage{needspace}

\newsavebox{\ablTopBox}

\title{When Your Agent Opens the Chat App: Agent-Controlled Search over Raw Chat Logs Rivals Structured Memory}

\author[1,*]{Ruizhe Li}
\author[1,\dagger]{Licheng Zhang}
\author[1,2]{Benfeng Xu}
\author[1]{Mingxuan Du}
\author[1]{Zheren Fu}
\author[1]{Weidong Chen}
\affiliation[1]{University of Science and Technology of China}
\affiliation[2]{MetaStone Technology}
\contribution[*]{Work done during an internship at MetaStone Technology.}
\contribution[\dagger]{Corresponding author.}
\abstract{%
Agent-memory systems increasingly buy retrieval quality with structure, transforming raw conversation histories into summaries, embeddings, trees, or knowledge graphs before any question is asked. We ask how much of that benefit comes from the structure itself, rather than from competent retrieval over the raw history. We present \textsc{ReFind}, an agent-controlled search interface that builds no semantic structure at all: it leaves the conversation archive unmodified, indexes it lexically at turn granularity, and combines a generic iterative keyword-search loop with four chat-native controls grounded in empirical refinding work: session-aware rank fusion, local context expansion, temporal narrowing, and skipping already-inspected sessions. A separate reasoning stage answers from the collected evidence. Across a broad suite of conversational-memory tasks (single- and multi-hop QA, event ordering, and fact consolidation), roughly 2{,}800 questions on precise-retrieval and fact-tracking capabilities evaluated under the incremental multi-turn setting of MemoryAgentBench, \textsc{ReFind} attains the highest mean accuracy (58.2) of any system compared, above the strongest graph- and tree-based memory systems (HippoRAG 2, 53.2), all under a GPT-4o-mini backbone matched to every reused baseline. Controlled comparisons to single-shot BM25, a matched generic-agentic BM25 control, component removals, and agentic dense/hybrid variants separately support the roles of agent control, chat-native controls, and lexical retrieval. On LongMemEval-S/M, the same interface reaches $93.2\pm3.3$ and $89.3\pm6.0$ with GPT-5-mini. The results indicate that for precise, evidence-grounded questions over chat archives, much of the benefit credited to elaborate memory structures is recoverable by giving an agent controllable search over the unmodified record, with no LLM-based index construction at all.
}
\date{August 16, 2026}

\begin{document}
\maketitle

\section{Introduction}
\label{sec:intro}
Large language model (LLM) agents have become a core paradigm for AI applications, demonstrating strong capabilities in tool use, code generation, scientific problem solving, and dialogue simulation \citep{yao2022react, shinn2023reflexion, wang2023voyager, park2023generative}. However, when agents operate over long-term tasks and multi-turn interactions, they face a fundamental bottleneck: \textit{memory} \citep{packer2023memgpt, zhang2024memorysurvey, yan2025gam}. As interaction histories grow beyond the context window, the system must decide what information to keep, compress, retrieve, or discard \citep{liu2024lost, hsieh2024ruler, hu2025mabench}.

The dominant response has been to impose structure \emph{ahead of time}. Agent-memory systems transform raw interaction histories into self-written memos, recursive summaries, vector stores, trees, or knowledge graphs \citep{lu2023memochat, wang2023recursively, chhikara2025mem0, sarthi2024raptor, edge2024graphrag, gutierrez2024hipporag}. Every such transformation is a bet placed before the question is known: it decides what to abstract and what to discard, and details omitted during preprocessing may be difficult to recover later \citep{packer2023memgpt, gutierrez2025hipporag2, yang2026stitch, yan2025gam}. It also carries a real cost, since indices must be built offline and rebuilt or incrementally maintained as new messages arrive.

This raises a question that the steady accumulation of increasingly elaborate memory architectures has left largely untested: \textit{how much of their reported benefit comes from the structure itself, rather than from having competent retrieval over the history at all?} To probe it, we take the opposite extreme---build no semantic structure whatsoever---and ask how far that setting can be pushed.

Our answer is grounded in empirical studies of personal-information refinding. Naturalistic observations show that people often reach known targets through small, context-guided steps rather than one fully specified keyword query \citep{teevan2004orienteering}. A direct instant-messaging study finds that users combine keyword trials with scrolling and temporal or contextual landmarks \citep{cheng2022reappropriation}; email studies likewise connect successful refinding to threading, local browsing, time and source cues, and search-history state \citep{whittaker2011organizing, elsweiler2011difficult, komlodi2007history}. This suggests preserving the raw record and moving intelligence from a pre-built semantic index into adaptive control over search.

We operationalize this principle as two layers: a \emph{generic agentic loop} that issues keyword queries, inspects results, saves evidence, and reformulates; and four \emph{chat-native controls} over where evidence lies in a conversational record---local context expansion, session-aware rank fusion, temporal narrowing, and seen-session filtering. This is what the title names: the agent \emph{opens the chat app} and tracks the answer down as a person would, rather than consulting a memory built for it in advance. The refinding studies motivate these four design dimensions, and our controlled ablations test their functional contribution in an agent search system.

The novelty lies in this system-level composition and in establishing what it buys relative to both structured memory and generic agentic retrieval. Across a broad suite of conversational-memory benchmarks---single- and multi-hop QA, event ordering, and fact consolidation, some 2{,}800 questions evaluated under the incremental multi-turn setting of MemoryAgentBench \citep{hu2025mabench}---it attains the highest mean accuracy of any system compared, above graph-based (GraphRAG, HippoRAG 2) and tree-based (RAPTOR) memory alike, all under a GPT-4o-mini backbone matched to every reused baseline. On the LongMemEval-S/M subsets \citep{wu2024longmemeval}, the advantage is reproduced across five runs with a stronger GPT-5-mini controller. The interface thereby replaces LLM-based offline index construction with question-directed online computation.

Our contributions are: (1) \textsc{ReFind}, an agent-controllable search interface for temporally ordered, session-structured conversational memory, combining lexical retrieval with session-aware reranking, context expansion, temporal filtering, and seen-session deduplication, driven by a ReAct agent that collects evidence before a separate answer-generation stage; (2) empirical evidence that on precise, evidence-grounded conversational-memory questions, much of the benefit credited to elaborate memory structures is recoverable without building any structure at all, and without any LLM-based index construction, provided the agent controls the search; and (3) a delineation of the interface's scope---effective for precise retrieval and factual-update tracking, complementary to semantic and low-latency memory mechanisms.

\section{Related Work}
\label{sec:related}
\paragraph{Agent Memory Systems.}
Human memory research distinguishes internal memory from external artifacts used for cognitive offloading \citep{clark1998extended, risko2016cognitive}, and agent memory systems face a similar choice: compress into internal or semi-structured memory, or preserve external records and retrieve from them. Retrieval-augmented generation \citep{lewis2020rag} retrieves external evidence at inference time; Mem0 \citep{chhikara2025mem0} extracts structured entries and stores them as embeddings; A-Mem \citep{xu2025amem} organizes memories as linked note cards; GraphRAG \citep{edge2024graphrag}, HippoRAG \citep{gutierrez2024hipporag, gutierrez2025hipporag2}, and Zep \citep{rasmussen2025zep} build graph or temporal structures for multi-hop reasoning; RAPTOR \citep{sarthi2024raptor} builds hierarchical trees by recursive summarization. Other systems vary the memory substrate: SeCom \citep{pan2025secom} compresses dialogue into topical segments, MemoRAG \citep{qian2024memorag} trains a memory model to generate retrieval clues, and MIRIX \citep{wang2025mirix} coordinates specialized memory agents. These methods gain efficiency and abstraction by preprocessing conversations into compact representations, at the cost of an offline construction pass that must be repeated or incrementally maintained as the archive grows. We study the complementary setting: preserving complete raw records and using search to recover precise details when compression may lose information.

\begin{table}[t]
\centering
\scriptsize
\setlength{\tabcolsep}{3pt}
\begin{tabular}{lccccc}
\toprule
\textbf{Method} & \textbf{No prep.} & \textbf{Agentic} & \textbf{Sess.} & \textbf{Time} & \textbf{Dedup} \\
\midrule
Sparse / dense RAG & \checkmark & & & & \\
RAPTOR, GraphRAG, & & & & & \\
\quad HippoRAG 2, A-Mem, & & & & & \\
\quad Mem0, STITCH & & & & & \\
Zep & & & & \checkmark & \\
SeCom & & & \checkmark & & \\
MemGPT, GAM & \checkmark & \checkmark & & & \\
\midrule
\textbf{\textsc{ReFind}} & \checkmark & \checkmark & \checkmark & \checkmark & \checkmark \\
\bottomrule
\end{tabular}
\caption{Design axes of conversational-memory systems. \textbf{No prep.}: no LLM-based offline index construction before questions arrive. \textbf{Agentic}: the model issues its own multi-round queries. \textbf{Sess.}: session structure informs ranking. \textbf{Time}: explicit temporal filtering as a retrieval control. \textbf{Dedup}: redundancy across search rounds is suppressed. No single axis is novel; the combination is what we study.}
\label{tab:design_axes}
\end{table}

\paragraph{Personal-Information Refinding.}
User studies provide a behavioral rationale for these controls. In a naturalistic diary and observation study, participants used keyword search in only 39\% of searches, more often ``orienteering'' through small local steps whose context informed the next action \citep{teevan2004orienteering}. A study of instant-messaging use (81 survey responses and 14 interviews) reports keyword trials, scrolling, and time-, visual-, and context-based landmarks when revisiting conversations \citep{cheng2022reappropriation}. At larger scale, a field study of 345 email users and over 85{,}000 refinding actions found search and threading more effective than complex preparatory foldering, with scrolling the most prevalent behavior \citep{whittaker2011organizing}. Smaller studies further link email-refinding difficulty to elapsed time and remembered temporal or source cues \citep{elsweiler2011difficult}, and show that search histories support progress tracking and task management \citep{komlodi2007history}. These findings motivate locality, hierarchy, chronology, and cross-round state; our ablations test how each becomes useful when implemented as an agent-facing retrieval control.

\paragraph{Agentic Retrieval.}
Several agent systems give models control over retrieval: MemGPT \citep{packer2023memgpt} pages information between context and external storage; Self-RAG \citep{asai2023selfrag} learns adaptive retrieval decisions; A-RAG \citep{du2026arag} exposes keyword, semantic, and chunk-reading tools for general document retrieval rather than conversational memory; GAM \citep{yan2025gam} preserves raw messages in page-stores and lets a Researcher Agent search over multiple rounds. \textsc{ReFind}'s defining combination is a generic iterative loop plus controls specialized to conversational structure: local expansion around matched turns, temporal constraints, session-level rank fusion, and explicit tracking of previously inspected sessions. Table~\ref{tab:design_axes} places these axes side by side.

\paragraph{Memory Evaluation Benchmarks.}
LongMemEval \citep{wu2024longmemeval} evaluates long-term memory through needle-in-a-haystack tasks across five dimensions; MemoryAgentBench \citep{hu2025mabench} recasts long-context datasets into incremental multi-turn formats, from which we draw our precise-retrieval and factual-update benchmarks; LoCoMo \citep{maharana2024locomo} constructs ultra-long 300--600 turn dialogues. We build our evaluation suite from the first two.

\section{Method}
\label{sec:method}
Our system combines generic iterative retrieval with controls tailored to text-only, timestamped, session-structured chat histories. Given a user's extensive conversation history and a question, it generates an answer through two decoupled stages: \textit{retrieval} (iterative evidence collection) and \textit{reasoning} (answer generation from collected evidence).

\paragraph{What ``no structure'' means.} \textsc{ReFind} uses the organization already present in a chat archive---turn boundaries, session identifiers, timestamps, and raw text---but creates no learned or LLM-generated memory representation. The BM25 index is an access path over the original text, not a surrogate memory: adding a message appends its tokens and metadata without summarizing earlier content, extracting entities, linking a graph, or rewriting a memory card. This distinction makes the comparison operational. Structured systems decide how to represent a record before seeing the future question; \textsc{ReFind} defers that decision until query time, when the agent knows what evidence it needs.

\subsection{System Architecture}

\begin{figure*}[t!]
\centering
\includegraphics[width=\textwidth]{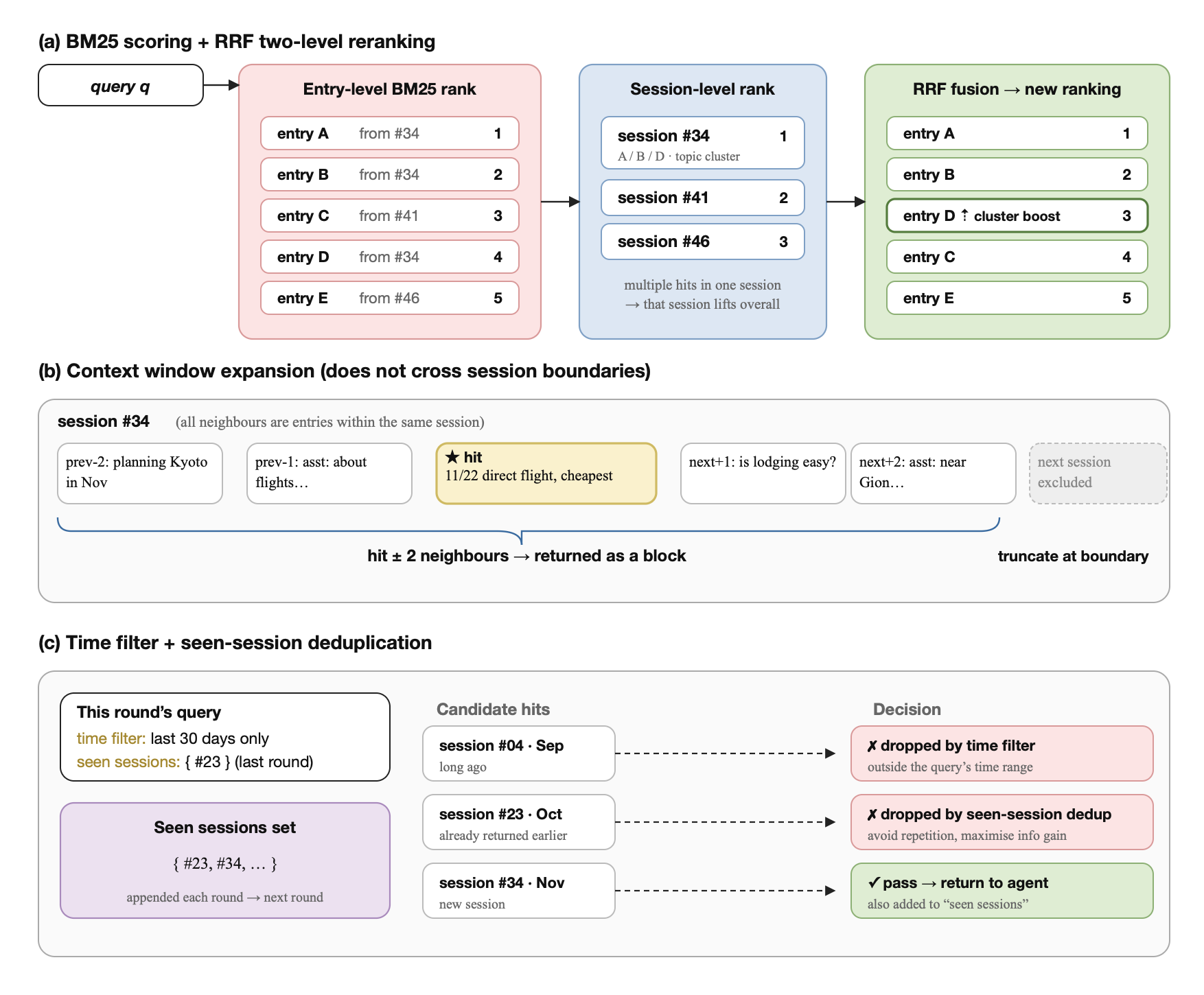}
\caption{Search engine core mechanisms. (a) BM25 scoring with RRF two-level reranking: turn-level BM25 ranks (shown as \textit{entries}) are fused with session-level aggregate ranks via Reciprocal Rank Fusion---multiple hits within the same session boost that session's entries. (b) Context window expansion: the matched entry plus $\pm$2 neighboring turns are returned as a block, truncated at session boundaries. (c) Temporal filtering and seen-session deduplication: candidates outside the time range or from previously returned sessions are dropped, maximizing information gain per iteration.}
\label{fig:search_engine}
\end{figure*}

In \textbf{Stage 1 (retrieval)}, a ReAct-style \citep{yao2022react} agent loop lets the LLM autonomously decide search keywords and parameters, conducting multiple rounds of search over the history with the search engine described below (Figure~\ref{fig:search_engine}) and saving valuable fragments as notes. A round can exploit the previous observation: the agent may replace an unsuccessful phrase with a distinctive entity, narrow a date range after finding a temporal cue, or search for a second fact needed by a multi-hop question. In \textbf{Stage 2 (reasoning)}, the collected notes are grouped by session, sorted chronologically, and presented with the question for answer generation. This separation prevents answer generation from competing with retrieval for context space: the search loop explores the archive, while the answer model receives a compact, evidence-only view in the archive's original wording.

\subsection{Search Engine Design}

Each component below operationalizes one chat-history axis motivated by the refinding studies above; the system is not intended as a cognitive model.

\subsubsection{Data Representation and Scoring}

Conversation histories exhibit a two-level structure: the finest unit is a \textit{turn}---a user utterance paired with its assistant response---and turns group into \textit{sessions}, complete dialogue episodes carrying a unique identifier and timestamp.

We build a BM25 inverted index \citep{robertson2009bm25} at turn granularity, tokenizing by lowercasing, whitespace and punctuation splitting, Porter stemming, and stopword removal, and score with $k_1 = 1.2$, $b = 0.75$. Indexing requires no model call and no summarization pass, so an archive is searchable as soon as it is written and new turns are appended incrementally. We choose sparse retrieval partly for this reason and partly because lexical matches are interpretable: the agent observes which terms succeeded or failed, a stronger feedback signal for reformulation than a similarity score.

\subsubsection{RRF Two-Level Reranking}

BM25 performs exact matching at the turn level but ignores session-level topic clustering signals: when multiple turns within the same session are matched, the session as a whole is likely highly relevant to the query. Inspired by passage-level evidence aggregation \citep{callan1994passage}, we employ Reciprocal Rank Fusion (RRF) \citep{cormack2009rrf} to combine two granularity levels of ranking.

For each turn $c$, we compute two rankings:

\textbf{First ranking} $r_1(c)$: Direct BM25 ranking of all turns by score.

\textbf{Second ranking} $r_2(c)$: Session-level aggregation. We first compute each session $s$'s aggregate score:
\begin{equation}
\text{Score}_{\text{session}}(s) = \sum_{c \in s} \text{BM25}(q, c)
\end{equation}
Sessions are ranked by aggregate score, and each turn inherits its session's rank: $r_2(c) = \text{rank}(\text{session}(c))$.

The final fused score is:
\begin{equation}
\text{RRF}(c) = \frac{1}{k + r_1(c)} + \frac{1}{k + r_2(c)}
\end{equation}
with $k = 60$. The system returns the top-$K$ turns by RRF score (default $K = 5$).

The intuition is: if multiple turns within a session match the query, even turns with individually low BM25 scores in that session may contain valuable contextual information and should receive a ranking boost.

\subsubsection{Context, Time, and Redundancy Controls}

\paragraph{Context window expansion} \textit{(after finding a relevant message, inspect its surroundings).} For each retrieved turn $c_i$ the system returns $w$ turns before and after (default $w=2$): $\{c_{i-w}, \ldots, c_i, \ldots, c_{i+w}\}$, truncated at session boundaries. This gives the agent dialogue context around each hit rather than an isolated lexical match.

\paragraph{Temporal filtering} \textit{(narrow to a remembered time range).} When a question involves a time period (e.g., ``last month''), the agent supplies start and end parameters, and turns are filtered by timestamp before BM25 scoring.

\paragraph{Session deduplication} \textit{(skip what has already been inspected).} The system maintains the set of sessions returned in prior rounds and excludes them from subsequent searches. Filtering is at the session rather than turn level, since turns within a session typically share a topic, so once a session has been seen, its remaining turns are unlikely to add information. This is the only operation that makes the interface stateful across rounds.

\paragraph{Retrieval as a search decision process.} The components above combine into a single state-dependent action space. At round $t$ the agent holds a state $\sigma_t=(\mathcal{N}_t,\mathcal{S}_t,\mathcal{O}_t)$---saved notes, sessions already returned, and the last observation---and acts by issuing a \emph{parameterized query} $a_t=(\mathbf{k}_t,\tau_t,K_t)$ of keywords, an optional time range, and a result budget, obtaining $\mathcal{O}_{t+1}=\Pi_w(\mathrm{RRF}(\mathrm{BM25}(\mathbf{k}_t,\mathcal{H}\mid\tau_t,\mathcal{S}_t)),K_t)$ over the raw archive $\mathcal{H}$. Conditioning on $\tau_t,\mathcal{S}_t$ drops temporally irrelevant and already-inspected candidates before scoring, and $\Pi_w$ expands each surviving hit by $\pm w$ neighbouring turns. Conventional RAG is the degenerate case $T=1$ with $\tau,\mathcal{S},\Pi_w$ inert; what the interface adds is thus not a better scoring function but this state-dependent action space, which is what the ablations isolate.

\section{Experiments}
\label{sec:experiments}
\subsection{Evaluation Setup}

\paragraph{Task suite.} Our evaluation uses the incremental multi-turn suite of MemoryAgentBench \citep{hu2025mabench}, which converts long-context datasets into a format that simulates an agent gradually accumulating history. We report the six benchmarks that probe the two capabilities central to archival chat search---precise retrieval and factual-update tracking: single-/multi-hop QA \citep{hsieh2024ruler}, LongMemEval \citep{wu2024longmemeval}, EventQA, and single-/multi-hop FactConsolidation (the last two introduced by MemoryAgentBench)---each under its official metric, 2{,}800 questions in total. The four retrieval benchmarks test whether a system can recover a specific piece of evidence from a long record; the two consolidation benchmarks test whether it can resolve multiple versions of a fact and return the current one. The overall score is the unweighted macro-average across the six benchmarks.

\subsection{Implementation Details}

The retrieval stage uses a ReAct framework with three tools: \texttt{search\_chatrecord} (BM25 search with keyword and time-range parameters), \texttt{take\_note} (save fragments), and \texttt{finish\_search}. The system prompt instructs the LLM to collect evidence only, review all results after each search, save relevant entries, and try diverse keywords. Retrieval runs at most 4 iterations with BM25 top-$K=5$, a $\pm$2-turn context window, and RRF session fusion. The fixed budget bounds online work while leaving the controller free to stop early when evidence is complete. The reasoning stage formats notes by session in chronological order and presents them with the question for answer generation. Consequently, retrieval decisions are visible in the saved-note trace and answers can be checked against verbatim turns rather than latent memory states. Full prompts and hyperparameters are in the supplement.

For the task suite, both our retrieval controller and answer model use GPT-4o-mini, matching the backbone of the reused baselines, and we evaluate under the same configuration as \citet{hu2025mabench}: the LME subtask uses GPT-4o as judge, and the remaining tasks use automatic metrics (SubEM, ROUGE-L, Recall@5). The suite comparison measures the end-to-end gain of the complete \textsc{ReFind} interface; the matched controls in Table~\ref{tab:ablation} then separate multi-round control, chat-native retrieval controls, and backend choice. For the backbone-scaling subsets we use GPT-5-mini and follow the STITCH evaluation configuration \citep{yang2026stitch}---the fixed subsets, GPT-4.1-mini judge, judge prompt, decoding settings, and verdict parsing---with all STITCH baselines and our GAM run sharing the GPT-5-mini generator.

\subsection{Baselines}

Baselines are listed with citations in Tables~\ref{tab:mabench_detail} and~\ref{tab:longmemeval}, spanning long-context models, sparse and dense RAG (including Contriever \citep{izacard2021contriever}), structured memory systems (tree-, graph-, and note-based), and agentic memory systems. For the task suite, values are reused from \citet{hu2025mabench}. On the backbone-scaling subsets we adopt the baseline set of STITCH \citep{yang2026stitch}; values are reused from that paper rather than reproduced locally, except GAM \citep{yan2025gam}, which we run ourselves and re-evaluate with the same GPT-4.1-mini judge used for our predictions.

\begin{table*}[t]
\centering
\scriptsize
\setlength{\tabcolsep}{4.5pt}
\begin{tabular}{llccccccc}
\toprule
\textbf{Category} & \textbf{Method} & SH-QA & MH-QA & LME & EventQA & FC-SH & FC-MH & \textbf{Avg} \\
\midrule
Long-Context & GPT-4o-mini & 64.0 & 43.0 & 30.7 & 59.0 & 45.0 & 5.0 & 41.1 \\
\midrule
\multirow{5}{*}{Sparse / Dense RAG}
& BM25-RAG & 66.0 & 56.0 & 45.3 & \textbf{74.6} & 48.0 & 3.0 & 48.8 \\
& Contriever & 22.0 & 31.0 & 15.7 & 66.8 & 18.0 & \underline{7.0} & 26.8 \\
& text-embed-3-small & 60.0 & 44.0 & 48.3 & 63.0 & 28.0 & 3.0 & 41.1 \\
& text-embed-3-large & 54.0 & 44.0 & 50.3 & 70.0 & 28.0 & 4.0 & 41.7 \\
& Qwen3-Embed-4B & 57.0 & 47.0 & 43.3 & 71.4 & 29.0 & 3.0 & 41.8 \\
\midrule
\multirow{6}{*}{Structured Memory}
& RAPTOR & 29.0 & 38.0 & 34.3 & 45.8 & 14.0 & 1.0 & 27.0 \\
& GraphRAG & 47.0 & 47.0 & 35.0 & 34.4 & 14.0 & 2.0 & 29.9 \\
& MemoRAG & 29.0 & 33.0 & 20.0 & 56.0 & 21.0 & \underline{7.0} & 27.7 \\
& HippoRAG 2 & \underline{76.0} & \underline{66.0} & \underline{50.7} & 67.6 & \underline{54.0} & 5.0 & \underline{53.2} \\
& Mem0 & 25.0 & 32.0 & 36.0 & 37.5 & 18.0 & 2.0 & 25.1 \\
& Zep & 44.0 & 25.0 & 38.3 & 42.5 & 7.0 & 3.0 & 26.6 \\
\midrule
\multirow{3}{*}{Agentic Memory}
& Self-RAG & 35.0 & 42.0 & 25.7 & 31.8 & 19.0 & 3.0 & 26.1 \\
& MemGPT & 41.0 & 38.0 & 32.0 & 26.2 & 28.0 & 3.0 & 28.0 \\
& MIRIX & 62.0 & 61.0 & 37.3 & 29.8 & 14.0 & 2.0 & 34.4 \\
\midrule
& \textbf{\textsc{ReFind}} & \textbf{83.0} & \textbf{69.0} & \textbf{51.3} & \underline{74.1} & \textbf{62.7} & \textbf{8.8} & \textbf{58.2} \\
\bottomrule
\end{tabular}
\caption{Per-benchmark accuracy (\%) across the full compared field on six conversational-memory benchmarks, GPT-4o-mini backbone matched across all systems; baseline numbers reused from \citet{hu2025mabench}. Per column, best in \textbf{bold} and second-best \underline{underlined}. Metric is SubEM except LME (GPT-4o LLM-as-judge). Sample sizes: SH-QA/MH-QA 100, LME 300, EventQA 1{,}500, FC-SH/FC-MH 400. SH-QA, MH-QA, LME, and EventQA probe precise retrieval; FC-SH and FC-MH probe factual-update tracking. \textbf{Avg} is the unweighted macro-average across the six benchmarks.}
\label{tab:mabench_detail}
\end{table*}

\subsection{Main Results: Precise Retrieval and Fact Tracking}

Table~\ref{tab:mabench_detail} reports the full compared field on six conversational-memory benchmarks, all under a GPT-4o-mini backbone matched across systems. The six fall into two task types: SH-QA, MH-QA, LME, and EventQA probe \emph{precise retrieval}---locating a specific fact buried in a long history---while FC-SH and FC-MH probe \emph{factual-update tracking}, requiring the latest value after an earlier fact has been overwritten. \textsc{ReFind} tops the six-benchmark leaderboard with a mean accuracy of 58.2, ahead of the strongest structured baseline HippoRAG 2 (53.2) and single-shot BM25-RAG (48.8), over 2{,}800 questions spanning both task types.

Per benchmark, \textsc{ReFind} is best on five of the six: SH-QA 83.0, MH-QA 69.0, LME 51.3, FC-SH 62.7, and FC-MH 8.8, with HippoRAG 2 the runner-up on four of them. On EventQA, \textsc{ReFind} and BM25-RAG are closely matched (74.1 and 74.6), showing that a strong lexical hit can already solve many single-event questions. The larger gains occur where one query is least likely to suffice: relative to BM25-RAG, \textsc{ReFind} adds $17$ points on SH-QA and $13$ on MH-QA, while also lifting both factual-consolidation tasks. This pattern connects the aggregate score to the mechanism: iterative reformulation helps assemble dispersed evidence, and chronology- and session-aware controls help distinguish the relevant version of a fact. The small-subset experiments below decompose these gains with matched agent and retrieval controls. All of this holds under GPT-4o-mini, showing that effective query formulation does not require a frontier model.

\section{Analysis and Discussion}
\label{sec:ablation}

\subsection{Backbone Scaling}

We next test whether the advantage persists with a stronger controller. We run \textsc{ReFind} on the LongMemEval-S (50 questions, $\sim$115k tokens each) and LongMemEval-M (15 questions, $\sim$500k tokens each) subsets used by STITCH \citep{yang2026stitch}, using GPT-5-mini for both retrieval control and answer generation. The comparison reuses STITCH's GPT-5-mini baselines and evaluation protocol, and we repeat \textsc{ReFind} five times to measure execution stability on the fixed question sets.

\begin{table}[t]
\centering
\small
\setlength{\tabcolsep}{3.8pt}
\begin{tabular}{lcc}
\toprule
\textbf{Method} & \textbf{S} & \textbf{M} \\
\midrule
GPT-5-mini (long-context) & 82.0 & 53.3 \\
text-embed-3-large (RAG) & 80.0 & 26.7 \\
\midrule
RAPTOR~\citep{sarthi2024raptor} & 48.0 & 46.7 \\
GraphRAG~\citep{edge2024graphrag} & 84.0 & 66.7 \\
HippoRAG 2~\citep{gutierrez2025hipporag2} & 80.0 & 66.7 \\
A-Mem~\citep{xu2025amem} & 74.0 & 66.7 \\
STITCH~\citep{yang2026stitch} & 86.0 & 80.0 \\
GAM~\citep{yan2025gam} & 70.0 & 60.0 \\
\midrule
\textbf{\textsc{ReFind} (5 runs)} & $\mathbf{93.2 \pm 3.3}$ & $\mathbf{89.3 \pm 6.0}$ \\
\bottomrule
\end{tabular}
\caption{Backbone-scaling observation on the LongMemEval-S/M subsets (accuracy \%), GPT-5-mini backbone. Representative baselines reused from STITCH \citep{yang2026stitch} (full set therein); GAM re-evaluated with our GPT-4.1-mini judge, and all baselines share the GPT-5-mini generator. For \textsc{ReFind}: mean $\pm$ sample std.\ over five uniformly judged runs on the fixed S (50 questions) and M (15 questions) subsets, quantifying execution variability rather than question-sampling uncertainty.}
\label{tab:longmemeval}
\end{table}

Across five runs uniformly judged by GPT-4.1-mini, \textsc{ReFind} reaches $93.2 \pm 3.3$ (S) and $89.3 \pm 6.0$ (M), above every backbone-matched structured system reused from STITCH---trees, graphs, note cards, learned segmentation---including STITCH (86.0/80.0). The closest design comparison is GAM (70.0/60.0), which also lets an agent search the raw record but lacks the complete session-, time-, context-, and redundancy-aware interface; the supplement traces one \textsc{ReFind} trajectory. Together with the GPT-4o-mini suite, these results show that the advantage holds across two controllers and two evaluation settings. The LongMemEval margin over HippoRAG 2 grows from 0.6 points in the GPT-4o-mini suite to 13.2 (S) and 22.6 (M) under the GPT-5-mini setting.

\subsection{Ablation Study}

The top panel of Table~\ref{tab:ablation} reports a matched generic-agentic control and three-run component ablations on the same LongMemEval-S/M subsets under GPT-5-mini, against the five-run full-method mean. Reporting repeated means and sample standard deviations exposes execution variability directly; the supplement lists every run and the exact configuration of each control.

\begin{table}[t]
\centering
\scriptsize
\setlength{\tabcolsep}{2pt}
\sbox{\ablTopBox}{\begin{tabular}{lcccc}
\toprule
\textbf{Variant} & \textbf{S} & \textbf{M} & \textbf{$\Delta$S} & \textbf{$\Delta$M} \\
\midrule
Full method (5 runs) & $93.2 \pm 3.3$ & $89.3 \pm 6.0$ & --- & --- \\
Generic agentic BM25 & $78.7 \pm 4.6$ & $82.2 \pm 3.8$ & $-$14.5 & $-$7.1 \\
\midrule
w/o context window & $84.0 \pm 0.0$ & $84.4 \pm 3.8$ & $-$9.2 & $-$4.9 \\
w/o session dedup. & $92.0 \pm 4.0$ & $80.0 \pm 6.7$ & $-$1.2 & $-$9.3 \\
w/o RRF reranking & $89.3 \pm 1.2$ & $84.4 \pm 7.7$ & $-$3.9 & $-$4.9 \\
w/o temporal filter & $91.3 \pm 4.2$ & $84.4 \pm 3.8$ & $-$1.9 & $-$4.9 \\
\midrule
One search (3 runs) & $84.7 \pm 3.1$ & $68.9 \pm 3.8$ & $-$8.5 & $-$20.4 \\
\bottomrule
\end{tabular}}%
\usebox{\ablTopBox}

\vspace{1pt}
\begin{tabular*}{\wd\ablTopBox}{@{\extracolsep{\fill}}lcc}
\toprule
\textbf{Backend} & \textbf{S} & \textbf{M} \\
\midrule
BM25 (5 runs) & $\mathbf{93.2 \pm 3.3}$ & $\mathbf{89.3 \pm 6.0}$ \\
Dense / emb.\ (3 runs) & $91.3 \pm 4.6$ & $82.2 \pm 7.7$ \\
Hybrid (3 runs) & $91.3 \pm 1.2$ & $86.7 \pm 0.0$ \\
\bottomrule
\end{tabular*}
\caption{Ablations on LongMemEval (accuracy \%; mean $\pm$ sample std.). \textbf{Top:} three-run generic-agentic, component, and one-search controls vs.\ the five-run full-method mean ($\Delta$ descriptive, not paired). \textbf{Bottom:} retrieval-backend comparison, agent/prompts/evaluation fixed.}
\label{tab:ablation}
\end{table}

\paragraph{Chat-native control layer.} To separate our interface from generic agentic RAG, we retain GPT-5-mini, the four-iteration ReAct controller, $K=5$, note-taking, and the answer stage, but make the tool return only turn-level BM25 hits: no local expansion, session RRF, temporal filtering, or seen-session filtering. This matched Generic Agentic BM25 control reaches $78.7 \pm 4.6$ (S) and $82.2 \pm 3.8$ (M), below the full method by 14.5 and 7.1 points. Iterative lexical search alone therefore does not explain \textsc{ReFind}'s result; the chat-native controls are jointly consequential under the fixed controller and budget.

\paragraph{Context window} removal is the largest S drop (Table~\ref{tab:ablation}): BM25-matched turns alone often lack sufficient information, and the surrounding dialogue is what makes a lexical hit interpretable. \textbf{RRF reranking} removal also costs several points, confirming that session-level topic clustering carries usable signal, while \textbf{temporal filtering} has the smallest S effect, consistent with the low proportion of time-sensitive questions in these subsets; its contribution appears instead in MemoryAgentBench selective forgetting, where tracking factual updates is the task.

\paragraph{Session deduplication} matters more on M ($-$9.3 points) than S ($-$1.2): without it, the longer M histories repeatedly return the same sessions and exhaust the iteration budget, supporting the value of state across rounds.

\paragraph{Search-loop control.} As a direct control for iteration, we let the unchanged controller formulate one BM25 query, force the reported $K=5$ and $\pm2$ context, pass every returned block to the unchanged answer stage, and forbid evidence selection or reformulation. This one-search variant (Table~\ref{tab:ablation}) trails the full method by 8.5 (S) and 20.4 (M) points---the multi-round loop's contribution with everything else fixed. The larger M gap is consistent with longer histories benefiting more from result-conditioned reformulation and skipping inspected sessions; the supplement reports per-run scores and resource use.

\paragraph{Retrieval backend.} Varying only the backend (agent, prompts, evaluation fixed), BM25 has the highest mean on both subsets; even Hybrid does not beat BM25's mean. This is consistent with the observation of \citet{wang2026bm25scale}, whose BM25+Dense RRF comparison likewise finds that fusion does not improve over BM25 alone. Adding dense signal thus does not raise accuracy here---lexical matching is not the bottleneck---and the supplement gives per-run detail.

\paragraph{What the controls establish.} The three control families rule out distinct simpler explanations. Generic Agentic BM25 keeps the multi-round controller but removes the four chat-native controls, so iteration alone does not recover the full score. One search keeps the chat-native retrieval behavior for the first query but removes reaction to returned evidence, so a well-formed initial query alone does not recover it either. Finally, the dense and four-way hybrid variants keep the controller and controls fixed while changing the retrieval signal; neither exceeds BM25. The resulting evidence chain attributes the gain to the interaction between multi-round control and a conversationally structured lexical interface, rather than to a larger model, semantic embeddings, or one unusually effective first query.

\subsection{Discussion}

\paragraph{Why lexical search remains competitive.} Chat archives contain many discriminative surface forms---names, products, places, dates, quoted phrases, and wording repeated across turns. A fixed dense query must encode the user's information need correctly in one representation, whereas the \textsc{ReFind} controller can observe a partial lexical match and convert newly exposed terms into the next query. Local expansion then restores pronouns and ellipses whose meaning lies in adjacent turns, while session fusion aggregates several weak hits into one coherent episode. This division of labor explains why BM25 leads the dense and hybrid backends in Table~\ref{tab:ablation}: the agent supplies semantic adaptation through reformulation, allowing the retrieval layer to remain exact, transparent, and incrementally maintainable.

\paragraph{Question-directed computation.} \textsc{ReFind} builds no LLM-generated index: its lexical index makes no model calls, and new turns are searchable immediately. Computation occurs only when a question arrives, averaging 2.5--2.6 searches and 5.0 LLM calls per query (supplement). By comparison, on one $\sim$1M-token corpus GraphRAG extracts 8{,}564 entities and 20{,}691 relations in 281 minutes of GPT-4-turbo time before answering a question \citep{edge2024graphrag}. \textsc{ReFind} therefore ties model computation to expressed information needs rather than to every archive update.

\paragraph{A memory-system design principle.} The results separate storage fidelity from access intelligence. Keeping raw history preserves every detail and makes updates append-only; agent control decides which details matter for the current question. Session, time, neighborhood, and visited-state metadata provide enough structure for the agent to traverse the archive without committing to a global semantic representation in advance. This suggests a modular default for conversational memory: begin with faithful storage and a controllable search interface, then add derived structures only for workloads that demand a separate latency or abstraction layer.

\paragraph{Coverage of chat-search failure modes.} A flat retriever can fail because a cue is underspecified, a hit lacks local context, evidence spans a session, the answer depends on time, or repeated queries revisit material. \textsc{ReFind} maps one mechanism to each: reformulation, local expansion, session fusion, temporal filtering, and deduplication. These mechanisms compose across rounds: a name exposed by one query can retrieve a session whose date constrains the next. The interface thus turns five static operations into a coherent evidence-gathering process.

\section{Conclusion}
\label{sec:conclusion}
We asked how much of an agent-memory system's benefit comes from pre-built structure, and answered by building none. \textsc{ReFind} leaves the archive unmodified and combines iterative lexical search with four chat-native controls. It tops the six-benchmark leaderboard at 58.2 mean accuracy (vs.\ 53.2 for the strongest structured baseline) over 2{,}800 questions with GPT-4o-mini, and reaches $93.2\pm3.3$/$89.3\pm6.0$ on LongMemEval-S/M with GPT-5-mini. Generic-agentic, component, one-search, and backend controls identify the effective design: iterative evidence collection over a session-aware, context-preserving, temporal, and deduplicated lexical interface.

The central result is that preserving raw records and exposing their conversational structure to an adaptive agent outperforms first transforming them into a memory representation. Question-directed search yields a strong, auditable, and immediately updatable memory system.

\section*{Statement on the Use of AI Assistants}
Claude Code and Codex assisted with code and editing; the authors verified all methods, experiments, and claims.

\bibliographystyle{assets/plainnat}
\bibliography{custom}

\clearpage
\beginappendix
\input{appendix}

\end{document}

%% file: appendix.tex
\section{Prompts}
\label{app:prompts}

\subsection{Stage 1: Retrieval Agent Prompt}

The retrieval agent uses the following system prompt:

\begin{mdframed}[linecolor=blue!45,backgroundcolor=blue!4,roundcorner=3pt,innertopmargin=4pt,innerbottommargin=4pt,innerleftmargin=6pt,innerrightmargin=6pt]
\begin{footnotesize}
\begin{verbatim}
You are a research assistant collecting
evidence from a user's conversation
history. You are NOT
answering — only gathering.

## Actions

search_chatrecord — Search by keywords.
Stemming enabled. Previously returned
sessions auto-excluded.
  {"keywords": ["keyword1", "keyword2"],
   "top_k": 5}

take_note — Save results from the LAST
search by number. Saves complete original
conversations.
  {"indices": [1, 3, 5]}

finish_search — Done collecting; proceed
to answer.
  {}

## Workflow
search → take_note → search again
→ ... → finish

## Rules
1. After EVERY search, review ALL results
   and save ANY that MIGHT be relevant.
   Missed info is LOST.
2. Call take_note BEFORE your next search.
   Previous results become inaccessible
   after a new search.
3. Don't answer — only collect. Try
   diverse keywords.
4. If no results are relevant, skip
   take_note and go directly to
   search_chatrecord or finish_search.

## Response Format

ONLY output Thought + Action + Action
Input. NEVER generate "Observation:" -
observations are provided by the system.
Keep Thought to 1-2 sentences.

Thought: Results 1, 3, and 7 mention
the topic. Saving them.
Action: take_note
Action Input: {"indices": [1, 3, 7]}
\end{verbatim}
\end{footnotesize}
\end{mdframed}

When temporal filtering is enabled, the following addendum is appended:

\needspace{12\baselineskip}
\begin{mdframed}[nobreak=true,linecolor=blue!45,backgroundcolor=blue!4,roundcorner=3pt,innertopmargin=4pt,innerbottommargin=4pt,innerleftmargin=6pt,innerrightmargin=6pt]
\begin{footnotesize}
\begin{verbatim}
## Time Filtering
Optional date range: add
date_from / date_to
(YYYY/MM/DD) to search_chatrecord.
  {"keywords": ["k1"],
   "date_from": "2023/06/01",
   "date_to": "2023/06/30"}

Broad first, narrow later: First search
WITHOUT time filters (catches
retrospective mentions). Then use time
filters to find
original events.
\end{verbatim}
\end{footnotesize}
\end{mdframed}

After each tool call, the following user message supplies the observation and requests the next action:

\needspace{10\baselineskip}
\begin{mdframed}[nobreak=true,linecolor=blue!45,backgroundcolor=blue!4,roundcorner=3pt,innertopmargin=4pt,innerbottommargin=4pt,innerleftmargin=6pt,innerrightmargin=6pt]
\begin{footnotesize}
\begin{verbatim}
Observation: {observation}

Respond with Thought + Action + Action
Input ONLY. Do NOT generate Observation
yourself.
\end{verbatim}
\end{footnotesize}
\end{mdframed}

\needspace{12\baselineskip}
\subsection{Stage 2: LongMemEval Reasoning Prompts}

\needspace{18\baselineskip}
\begin{mdframed}[nobreak=true,linecolor=blue!45,backgroundcolor=blue!4,roundcorner=3pt,innertopmargin=4pt,innerbottommargin=4pt,innerleftmargin=6pt,innerrightmargin=6pt]
\begin{footnotesize}
\begin{verbatim}
System: You are an AI assistant with
long-term memory. Answer questions based
on the user's conversation history. Read
all sessions carefully and extract
relevant information. If the answer cannot
be found,
clearly state insufficient information.
Requirements:
1. Answer directly and concisely
2. If time reasoning is needed, explain
   the process
3. Keep the answer brief and accurate

User: Below is the relevant conversation
history:
{notes}
---
Current time: {question_date}
Question: {question}
Answer based on the conversation history
above.
\end{verbatim}
\end{footnotesize}
\end{mdframed}

If Stage 1 saves no evidence, the same system message is paired with the following user template:

\needspace{13\baselineskip}
\begin{mdframed}[nobreak=true,linecolor=blue!45,backgroundcolor=blue!4,roundcorner=3pt,innertopmargin=4pt,innerbottommargin=4pt,innerleftmargin=6pt,innerrightmargin=6pt]
\begin{footnotesize}
\begin{verbatim}
User: No conversation history relevant to
the question was found.

---
Current time: {question_date}
Question: {question}

Answer based on the conversation history
above.
\end{verbatim}
\end{footnotesize}
\end{mdframed}

\subsection{Stage 2: MemoryAgentBench Task Prompts}

For the six MemoryAgentBench scores reported in the main paper, Stage 1 is unchanged and Stage 2 uses the following common system message:

\needspace{9\baselineskip}
\begin{mdframed}[nobreak=true,linecolor=blue!45,backgroundcolor=blue!4,roundcorner=3pt,innertopmargin=4pt,innerbottommargin=4pt,innerleftmargin=6pt,innerrightmargin=6pt]
\begin{footnotesize}
\begin{verbatim}
You are a helpful assistant that can read
the context and memorize it for future
retrieval.
\end{verbatim}
\end{footnotesize}
\end{mdframed}

The corresponding user templates are:

\needspace{18\baselineskip}
\begin{mdframed}[nobreak=true,linecolor=blue!45,backgroundcolor=blue!4,roundcorner=3pt,innertopmargin=4pt,innerbottommargin=4pt,innerleftmargin=6pt,innerrightmargin=6pt]
\begin{footnotesize}
\begin{verbatim}
Single-hop and multi-hop QA:
Below are the relevant documents retrieved
from memory:

{notes}
---
Answer the question based on the memorized
documents. Only give me the answer and do
not output any other words.

Now Answer the Question: {question}
\end{verbatim}
\end{footnotesize}
\end{mdframed}

\needspace{18\baselineskip}
\begin{mdframed}[nobreak=true,linecolor=blue!45,backgroundcolor=blue!4,roundcorner=3pt,innertopmargin=4pt,innerbottommargin=4pt,innerleftmargin=6pt,innerrightmargin=6pt]
\begin{footnotesize}
\begin{verbatim}
LongMemEval:
Below are the relevant chat history
retrieved from memory:

{notes}
---
The history chats are between you and a
user. Based on the relevant chat history,
answer the question as concisely as you
can, using a single phrase if possible.

{question}

Answer:
\end{verbatim}
\end{footnotesize}
\end{mdframed}

\needspace{16\baselineskip}
\begin{mdframed}[nobreak=true,linecolor=blue!45,backgroundcolor=blue!4,roundcorner=3pt,innertopmargin=4pt,innerbottommargin=4pt,innerleftmargin=6pt,innerrightmargin=6pt]
\begin{footnotesize}
\begin{verbatim}
EventQA:
Below are the relevant passages retrieved
from memory:

{notes}
---
Based on the context you memorized,
complete the task below:

{question}

The event that happens next is:
\end{verbatim}
\end{footnotesize}
\end{mdframed}

\needspace{30\baselineskip}
\begin{mdframed}[nobreak=true,linecolor=blue!45,backgroundcolor=blue!4,roundcorner=3pt,innertopmargin=4pt,innerbottommargin=4pt,innerleftmargin=6pt,innerrightmargin=6pt]
\begin{footnotesize}
\begin{verbatim}
Fact Consolidation:
Below are the relevant facts retrieved
from the knowledge pool:

{notes}
---
Pretend you are a knowledge management
system. Each fact in the knowledge pool is
provided with a serial number at the
beginning, and the newer fact has larger
serial number. You need to solve the
conflicts of facts in the knowledge pool
by finding the newest fact with larger
serial number. You should give a very
concise answer without saying other words
only from the knowledge pool you have
memorized rather than the real facts in
real world.

For example:
[Knowledge Pool]
Question: Based on the provided Knowledge
Pool, what is the name of the current
president of Russia?
Answer: Donald Trump

Now Answer the Question: Based on the
provided Knowledge Pool, {question}
Answer:
\end{verbatim}
\end{footnotesize}
\end{mdframed}

\Needspace{28\baselineskip}
\section{Hyperparameters}
\label{app:hyperparams}

\begin{table}[!h]
\centering
\small
\begin{tabular}{@{}l@{\hspace{8pt}}c@{}}
\toprule
\textbf{Parameter} & \textbf{Value} \\
\midrule
\multicolumn{2}{l}{\textit{BM25 Search Engine}} \\
$k_1$ (term frequency saturation) & 1.2 \\
$b$ (length normalization) & 0.75 \\
Top-$K$ per search & 5 \\
Context window & $\pm$2 conversations \\
RRF smoothing constant $k$ & 60 \\
\midrule
\multicolumn{2}{l}{\textit{ReAct Agent}} \\
Max iterations & 4 \\
Session deduplication & Enabled \\
Temporal filtering & Enabled \\
\midrule
\multicolumn{2}{l}{\textit{LLM Configuration}} \\
Backbone (LongMemEval) & GPT-5-mini \\
Backbone (MABench) & GPT-4o-mini \\
Temperature & 0 \\
Stage-1 max output tokens & 4096 \\
Stage-2 max output tokens & 4096 \\
\bottomrule
\end{tabular}
\caption{Hyperparameter settings.}
\label{tab:hyperparams}
\end{table}

\section{Evaluation Details}
\label{app:eval}

\subsection{Baseline Sources}

For LongMemEval, most baseline numbers are reused from STITCH \citep{yang2026stitch}, which reports results on the same LongMemEval-S and LongMemEval-M subsets used in our experiments. The backbone generator is matched: STITCH documents that all of its baseline methods use GPT-5-mini as the backbone LLM generator, and our controller and answer model also use GPT-5-mini. Our GAM run also uses GPT-5-mini as the backbone generator, and we re-evaluate its outputs with the same GPT-4.1-mini judge and verdict parsing protocol described below. A common judge and a common backbone improve comparability, but the systems were not all rerun with a matched controller or tool-call budget.

For MABench, baseline numbers are reused from the MABench paper \citep{hu2025mabench}. Our method is evaluated under the same benchmark setting, using GPT-4o-mini as the backbone to match the reported long-context and memory-system comparisons. The LME subtask uses the MABench LLM-as-judge protocol, while the remaining subtasks use the official automatic metrics. The full 16-system, six-benchmark comparison and its unweighted macro-average are reported in the main paper, where \textsc{ReFind} attains the highest mean accuracy (58.2).

\subsection{LongMemEval Evaluation}

We follow the STITCH evaluation protocol exactly:
\begin{itemize}
\item \textbf{Judge model:} gpt-4.1-mini (temperature=0.0, top\_p=0.9)
\item \textbf{Prompt:} ``Your task is to label an answer to a question as CORRECT or WRONG. You will be given: (1) a question, (2) a gold answer, and (3) a generated answer. The gold answer is concise and contains the ground-truth information. The generated answer might be longer. Be generous: if the generated answer contains the gold answer information (even verbatim inside a longer response), mark it as CORRECT. Otherwise, mark it as WRONG. Respond with either CORRECT or WRONG, and provide a brief reasoning.''
\item \textbf{Verdict parsing:} Response contains ``correct'' and does not contain ``wrong'' $\rightarrow$ CORRECT.
\end{itemize}

The retrieval and answer calls use temperature 0, but the external API does not expose a reproducible sampling seed. Repeated runs therefore quantify residual variation from model serving and agent trajectories. We did not conduct an independent human-agreement study for the automatic judge.

\subsection{Repeated-Run Stability}

We uniformly re-judge five full-method runs with GPT-4.1-mini. The runs include the full-method execution used to anchor the ablation study and four other executions. For the stability summary, the ablation-anchor execution is re-judged with the same judge as the other four runs. The main paper's ablation table retains its original paired evaluation and is interpreted only through within-evaluation deltas.

\begin{table}[tbp]
\centering
\small
\begin{tabular}{@{}l@{\hspace{8pt}}c@{\hspace{8pt}}c@{}}
\toprule
\textbf{Run} & \textbf{S} & \textbf{M} \\
\midrule
Run 1 & 96.0 & 86.7 \\
Run 2 (ablation anchor, re-judged) & 94.0 & 93.3 \\
Run 3 & 92.0 & 93.3 \\
Run 4 & 88.0 & 80.0 \\
Run 5 & 96.0 & 93.3 \\
\midrule
Mean $\pm$ sample std. & $93.2 \pm 3.3$ & $89.3 \pm 6.0$ \\
\bottomrule
\end{tabular}
\caption{Five-run LongMemEval stability on the same fixed subsets (accuracy \%). Repeated runs measure execution variability, not uncertainty from sampling questions.}
\label{tab:five_run_stability}
\end{table}

\subsection{Generic Agentic BM25 Control}
\label{app:generic_agentic}

This control isolates the chat-native interface from generic multi-round retrieval. It retains GPT-5-mini, the same four-iteration ReAct controller, $K=5$, BM25 tokenization, note-taking, and Phase-2 answer stage. The search tool returns only the matched turn and disables local context expansion, session-level RRF and neighbour propagation, temporal filtering, and seen-session filtering. Every tool call is audited to ensure that no removed control enters through an agent-supplied parameter.

\begin{table}[tbp]
\centering
\small
\begin{tabular}{@{}l@{\hspace{8pt}}c@{\hspace{8pt}}c@{\hspace{8pt}}c@{}}
\toprule
\textbf{Set} & \textbf{Run accuracies} & \textbf{Mean $\pm$ std.} & \textbf{$\Delta$ Full} \\
\midrule
S & 76.0 / 84.0 / 76.0 & $78.7 \pm 4.6$ & $-14.5$ \\
M & 80.0 / 86.7 / 80.0 & $82.2 \pm 3.8$ & $-7.1$ \\
\bottomrule
\end{tabular}
\caption{Three-run Generic Agentic BM25 control on the fixed LongMemEval subsets (accuracy \%; sample standard deviation). The delta compares its three-run mean with the five-run full-method mean.}
\label{tab:generic_agentic_runs}
\end{table}

Generic iterative BM25 does not recover the full method: jointly adding the four chat-native controls corresponds to 14.5 and 7.1 points on S and M, respectively, under the fixed controller and online budget. For context, Generic Agentic BM25 is below the one-search control on S (78.7 vs.\ 84.7) but above it on M (82.2 vs.\ 68.9). This crossover is descriptive rather than a component contrast because one-search retains all four chat-native retrieval controls; it is consistent with generic iteration sometimes revisiting noisy, uncontextualized hits on S while remaining useful on M's longer histories. M has only 15 questions, so the ordering requires caution.

\subsection{Repeated Component Ablations}
\label{app:component_ablation_runs}

For each component ablation, we retain the previously reported execution and add two independent GPT-5-mini executions evaluated with the same GPT-4.1-mini judge configuration. Each condition changes only the component named in Table~\ref{tab:component_ablation_runs}; all other prompts, budgets, and retrieval settings remain fixed.

\begin{table}[tbp]
\centering
\small
\begin{tabular}{@{}l@{\hspace{5pt}}l@{\hspace{5pt}}c@{\hspace{5pt}}c@{}}
\toprule
\textbf{Variant} & \textbf{Set} & \textbf{Run accuracies} & \textbf{Mean $\pm$ std.} \\
\midrule
\multirow{2}{*}{w/o context window} & S & 84.0 / 84.0 / 84.0 & $84.0 \pm 0.0$ \\
& M & 80.0 / 86.7 / 86.7 & $84.4 \pm 3.8$ \\
\midrule
\multirow{2}{*}{w/o session dedup.} & S & 88.0 / 96.0 / 92.0 & $92.0 \pm 4.0$ \\
& M & 73.3 / 80.0 / 86.7 & $80.0 \pm 6.7$ \\
\midrule
\multirow{2}{*}{w/o RRF reranking} & S & 90.0 / 90.0 / 88.0 & $89.3 \pm 1.2$ \\
& M & 80.0 / 80.0 / 93.3 & $84.4 \pm 7.7$ \\
\midrule
\multirow{2}{*}{w/o temporal filter} & S & 90.0 / 88.0 / 96.0 & $91.3 \pm 4.2$ \\
& M & 86.7 / 80.0 / 86.7 & $84.4 \pm 3.8$ \\
\bottomrule
\end{tabular}
\caption{Three-run component ablations on the fixed LongMemEval subsets (accuracy \%; sample standard deviation). Run 1 is the previously reported execution; runs 2 and 3 are new independent executions.}
\label{tab:component_ablation_runs}
\end{table}

Against the five-run full-method means (93.2 on S and 89.3 on M), the component-ablation mean differences are respectively $-9.2/-4.9$ (context window), $-1.2/-9.3$ (session deduplication), $-3.9/-4.9$ (RRF reranking), and $-1.9/-4.9$ (temporal filtering), reported as S/M. Context is the clearest S contribution, while session deduplication has the largest M mean effect. Because M contains only 15 questions, one changed judgment moves accuracy by 6.7 points; the run-level dispersion is therefore essential to interpreting the M ordering.

\subsection{One-Search Control}
\label{app:one_search}

This control preserves the full method's first-query intelligence while removing its ability to react to retrieval. The unchanged Phase-1 controller formulates one \texttt{search\_chatrecord} action; we force the reported BM25 configuration ($K=5$, context window $\pm2$, neighbour scoring enabled), then copy every returned result block into the unchanged Phase-2 answer context. There is no \texttt{take\_note} selection, second search, or query reformulation after observing the results. Thus, unlike raw-question BM25-RAG, the control retains the agent-written first query and isolates the value of subsequent rounds.

\begin{table}[tbp]
\centering
\small
\begin{tabular}{@{}l@{\hspace{8pt}}c@{\hspace{8pt}}c@{\hspace{8pt}}c@{}}
\toprule
\textbf{Method} & \textbf{Runs} & \textbf{S} & \textbf{M} \\
\midrule
Full multi-round \textsc{ReFind} & 5 & $93.2 \pm 3.3$ & $89.3 \pm 6.0$ \\
One search, all results & 3 & $84.7 \pm 3.1$ & $68.9 \pm 3.8$ \\
\midrule
Observed multi-round gain & --- & $+8.5$ & $+20.4$ \\
\bottomrule
\end{tabular}
\caption{Controlled search-loop ablation on the fixed LongMemEval subsets (accuracy \%; mean $\pm$ sample standard deviation). One-search run scores are 82.0/88.0/84.0 on S and 66.7/73.3/66.7 on M. Because repetition counts differ, gains are differences of observed means rather than paired estimates.}
\label{tab:one_search_control}
\end{table}

The full loop improves the observed mean by 8.5 points on S and 20.4 on M. This asymmetry is mechanistically consistent with M's longer archives: after inspecting an initial result, the full agent can preserve useful evidence, reformulate around newly exposed terms, and exclude already-inspected sessions. The control cannot determine which later-round operation contributes how much; the component ablations in the main paper address those operations separately.

\subsection{Retrieval-Backend Ablation}
\label{app:retrieval_backend}

We retain the same agent, prompts, four-iteration budget, $K=5$, controls, GPT-5-mini controller/answer model, and GPT-4.1-mini evaluator. With RRF $k=60$, \textbf{BM25} fuses turn rank and summed-BM25 session rank; \textbf{Dense} fuses \texttt{text-embedding-3-large} cosine rank and top-three-similarity session rank; the new four-way \textbf{Hybrid} fuses all four ranks.

\begin{table}[tbp]
\centering
\small
\begin{tabular}{@{}l@{\hspace{2.2pt}}l@{\hspace{2.2pt}}c@{\hspace{2.2pt}}c@{}}
\toprule
\textbf{Backend} & \textbf{Set} & \textbf{Run accuracies} & \textbf{Mean $\pm$ std.} \\
\midrule
\multirow{2}{*}{BM25 (5)} & S & 96.0 / 94.0 / 92.0 / 88.0 / 96.0 & $\mathbf{93.2 \pm 3.3}$ \\
& M & 86.7 / 93.3 / 93.3 / 80.0 / 93.3 & $\mathbf{89.3 \pm 6.0}$ \\
\midrule
\multirow{2}{*}{Dense (3)} & S & 94.0 / 86.0 / 94.0 & $91.3 \pm 4.6$ \\
& M & 86.7 / 86.7 / 73.3 & $82.2 \pm 7.7$ \\
\midrule
\multirow{2}{*}{Hybrid (4-way, 3)} & S & 90.0 / 92.0 / 92.0 & $91.3 \pm 1.2$ \\
& M & 86.7 / 86.7 / 86.7 & $86.7 \pm 0.0$ \\
\bottomrule
\end{tabular}
\caption{Matched retrieval-backend ablation on the fixed LongMemEval subsets (accuracy \%; sample standard deviation). BM25 has five runs; newer Dense and Hybrid conditions have three.}
\label{tab:retrieval_backend}
\end{table}

BM25 has the highest observed mean on both subsets. Hybrid ties Dense on S with lower variance and improves M by 4.4 points, but remains 1.9 (S) and 2.6 (M) points below BM25. M has only 15 questions, so one answer changes accuracy by 6.7 points; zero run-level variance does not imply deterministic outputs or a noiseless evaluator.

\subsection{MABench Evaluation}

For the LME subtask (the only subtask requiring LLM-as-judge), we follow the MABench evaluation protocol:
\begin{itemize}
\item \textbf{Judge model:} GPT-4o (temperature=0)
\item \textbf{Prompts:} Task-specific templates from the MABench codebase, including:
  \begin{itemize}
  \item Standard template for single-session and multi-session questions
  \item Temporal-reasoning template with off-by-one tolerance
  \item Knowledge-update template accepting updated answers alongside previous information
  \item Single-session-preference template using rubric-based evaluation
  \end{itemize}
\item \textbf{Verdict parsing:} Response contains ``yes'' $\rightarrow$ correct.
\end{itemize}

The remaining reported subtasks use SubEM (substring exact match).

\section{Resource Use}
\label{app:cost}

Table~\ref{tab:runtime_cost} reports resource use measured from logged agent trajectories and raw result files. For the full method, search and LLM-call counts are averages over all tasks in all five runs, and token ranges are the minimum and maximum of the five run-level mean token counts. The ablation-anchor run predates timing instrumentation. Full-method time therefore reports the initial timed execution followed by the mean of the three later timed reruns. Effective time sums successful LLM API time and retrieval time; it excludes local orchestration and retry waiting. Generic Agentic BM25 and one-search values are three-run means.

\begin{table}[tbp]
\centering
\small
\begin{tabular}{@{}l@{\hspace{3pt}}l@{\hspace{3pt}}c@{\hspace{3pt}}c@{\hspace{3pt}}c@{\hspace{3pt}}c@{}}
\toprule
\textbf{Method} & \textbf{Set} & \textbf{Searches} & \textbf{Calls} & \textbf{Tokens/task} & \textbf{Eff. time/task} \\
\midrule
\multirow{2}{*}{Full} & S & 2.61 & 4.99 & 69.8K--76.4K & 119.3s / 41.0s \\
& M & 2.43 & 4.99 & 83.1K--99.2K & 143.5s / 42.3s \\
\midrule
\multirow{2}{*}{Generic agentic} & S & 2.03 & 4.99 & 17.4K & 31.3s \\
& M & 2.20 & 4.98 & 17.0K & 31.9s \\
\midrule
\multirow{2}{*}{One search} & S & 1.00 & 2.00 & 9.6K & 14.5s \\
& M & 1.00 & 2.00 & 10.3K & 16.4s \\
\bottomrule
\end{tabular}
\caption{Average resource use. Full-method effective time reports the initial timed execution followed by the mean of three later timed reruns. Both use the same algorithmic settings; the difference reflects external API routing, queueing, service load, and serving state rather than a method change. Generic-agentic and one-search time is from contemporaneous three-run controls. Monetary charges were not recorded consistently, so we report token counts instead.}
\label{tab:runtime_cost}
\end{table}

\section{Artifact Use and Licenses}
\label{app:artifacts}

Our experiments use publicly available research artifacts, including LongMemEval \citep{wu2024longmemeval}, MemoryAgentBench \citep{hu2025mabench}, and the benchmark datasets incorporated by MemoryAgentBench, such as RULER \citep{hsieh2024ruler}, ReDial \citep{li2018conversational}, InfiniteBench \citep{zhang2024infinitebench}, and DetectiveQA \citep{xu2024detectiveqa}. We use these artifacts only for research evaluation, following the task definitions and evaluation protocols provided by their creators. We do not repurpose benchmark data for model training, user profiling, or deployment outside the evaluation setting.

We rely on the licenses and terms of use specified by the original artifact providers. Any code, prompts, processed outputs, and evaluation scripts released with our submission are intended to support research reproducibility and should be used consistently with the licenses and access conditions of the underlying benchmarks and model APIs.

\section{Ablation Variant Definitions}
\label{app:ablation}

\begin{itemize}
\item \textbf{Full method (baseline):} \\
RRF session fusion + context window ($\pm$2) + session deduplication + temporal filtering.
\item \textbf{Generic Agentic BM25:} \\
Retain the four-iteration ReAct loop, $K=5$, note-taking, and answer stage, but return only turn-level BM25 hits and jointly disable context expansion, session RRF, temporal filtering, and seen-session filtering.
\item \textbf{w/o context window:} \\
Set context\_window = 0. Only the directly matched conversation is returned, without surrounding context.
\item \textbf{w/o session dedup.:} \\
Disable session deduplication. Previously returned sessions can appear again in subsequent searches.
\item \textbf{w/o RRF reranking:} \\
Remove RRF session-level fusion. Use pure conversation-level BM25 ranking without session aggregation.
\item \textbf{w/o temporal filter:} \\
Disable the date\_from/date\_to parameters. The agent cannot restrict searches by time range.
\item \textbf{One search, all results:} \\
Use the unchanged controller to formulate the first query, execute one reported-configuration BM25 search, pass every returned block directly to the unchanged answer stage, and disable note selection and all later search rounds.
\end{itemize}

\section{Example Search Trajectory}
\label{app:trajectory}

Table~\ref{tab:trajectory_case} shows an example trajectory produced by our retrieval agent on a LongMemEval-S question. The question asks whether the user visited a museum with a friend ``two months ago.'' This case illustrates how the agent combines broad recall, note-taking, temporal filtering, and seen-session deduplication to disambiguate among several similar museum-related memories.

The trajectory is reproduced from raw execution logs. The \texttt{exclude\_sessions} field and the session identifiers shown in Step 3 are populated and consumed automatically by the search backend, which maintains the seen-session set as described in the main paper; they are internal log fields and are not part of the prompt or search results visible to the agent, which only issues keywords and optional date ranges.

\begin{table}[!h]
\centering
\small
\begin{tabular}{@{}p{0.05\columnwidth}@{\hspace{5pt}}p{0.43\columnwidth}@{\hspace{5pt}}p{0.44\columnwidth}@{}}
\toprule
\textbf{\#} & \textbf{Action and input} & \textbf{Outcome} \\
\midrule
1 &
Search, $K=10$: \texttt{museum}, \texttt{visited museum}, \texttt{went to museum}, \texttt{friend}, \texttt{went with}, \texttt{two months ago}. &
Returns several museum candidates: an Oct.\ 2022 Science Museum visit with a friend, a Feb.\ 2023 Natural History Museum visit with the user's dad, and the Jan.\ 2023 History Museum/Petra lecture. \\
\midrule
2 &
Save indices \texttt{[1,2,3,4,5,6,9]}. &
Preserves seven positive candidates and distractors before the next search. \\
\midrule
3 &
Search, $K=10$, Jan.\ 1--31, 2023: \texttt{museum}, \texttt{with my friend}, \texttt{with a friend}, \texttt{dad}, \texttt{visited}, \texttt{two months ago}. Exclude the six session IDs returned in Step~1. &
Maps ``two months ago'' from Mar.\ 11, 2023 to Jan.\ 2023 and excludes Step-1 sessions, surfacing new rather than repeated evidence. \\
\midrule
4 &
Save index \texttt{[1]}. &
Adds one Jan.\ 2023 candidate, for eight notes total. \\
\midrule
5 &
Reason over saved notes. &
Answers ``No'': the Jan.\ 11 History Museum/Petra memory has no friend; the friend and dad visits are temporally mismatched. \\
\bottomrule
\end{tabular}
\caption{Example retrieval trajectory for LongMemEval-S question \texttt{0bc8ad93}. Broad recall, temporal filtering, and seen-session deduplication disambiguate the target from similar but temporally mismatched memories.}
\label{tab:trajectory_case}
\end{table}

\paragraph{Saved evidence.}
The first note-taking step stores heterogeneous candidates: a Science Museum visit with a friend (Oct. 2022), several art or natural-history museum mentions, and the relevant History Museum lecture about ancient civilizations and Petra (Jan. 11, 2023). The second note-taking step adds a January candidate from a newly surfaced session. During reasoning, the agent compares these notes rather than relying on the top search hit alone. This is crucial because the strongest lexical distractor explicitly contains both ``museum'' and ``friend,'' but falls outside the intended time window.